\documentclass[]{spie}  

\usepackage{amsmath,amsfonts,amssymb}
\usepackage{graphicx}
\usepackage{xcolor}
\usepackage{cite}
\usepackage{algorithm}
\usepackage{algpseudocode}
\usepackage{booktabs}
\usepackage{multirow}

\title{Intrinsic Tolerance in C-Arm Imaging: How Extrinsic Re-optimization Preserves 3D Reconstruction Accuracy}

\author[a]{Lin Li}
\author[b]{Benjamin Aubert}
\author[a]{Paul Kemper}
\author[a]{Aric Plumley}

\affil[a]{AIX, Alphatec Spine, 1950 Camino Vida Roble, Carlsbad, CA 92008, USA}
\affil[b]{EOS, EOS Imaging Inc., 7275 Rue Saint-Urbain Suite 400, Montreal, QC H2R 2Y5, Canada}

\authorinfo{Correspondence: Lin Li, E-mail: linli@atecspine.com}

\begin{document}
\maketitle

\begin{abstract}
\textbf{Purpose:} C-arm fluoroscopy's 3D reconstruction relies on accurate intrinsic calibration, which is often challenging in clinical practice. This study ensures high-precision reconstruction accuracy by re-optimizing the extrinsic parameters to compensate for intrinsic calibration errors.

\noindent\textbf{Methods:} We conducted both simulation and real-world experiments using five commercial C-arm systems. Intrinsic parameters were perturbed in controlled increments. Focal length was increased by ±100 to ±700 pixels ($\approx$20 mm to 140 mm) and principal point by ±20 to ±200 pixels. For each perturbation, we (1) reconstructed 3D points from known phantom geometries, (2) re-estimated extrinsic poses using standard optimization, and (3) measured reconstruction and reprojection errors relative to ground truth.

\noindent\textbf{Results:} Even with focal length errors up to ±500 pixels ($\approx$100 mm, assuming a nominal focal length of $\sim$1000 mm), mean 3D reconstruction error remained under 0.2 mm. Larger focal length deviations (±700 pixels) elevated error to only $\approx$0.3 mm. Principal point shifts up to ±200 pixels introduced negligible reconstruction error once extrinsic parameters were re-optimized, with reprojection error increases below 0.5 pixels.

\noindent\textbf{Conclusion:} Moderate errors in intrinsic calibration can be effectively mitigated by extrinsic re-optimization, preserving submillimeter 3D reconstruction accuracy. This intrinsic tolerance suggests a practical pathway to relax calibration precision requirements, thereby simplifying C-arm system setup and reducing clinical workflow burden without compromising performance.
\end{abstract}


\section{INTRODUCTION}
\label{sec:intro}

C-arm fluoroscopy provides real-time X-ray views from multiple angles, enabling minimally invasive and image-guided surgical procedures such as orthopedic fixation, spinal instrumentation, and interventional radiology.\cite{feldman2007minimally, chen2015c} In quantitative applications like 3D reconstruction \cite{suh2025pedicle}, surgical navigation, and intraoperative registration, these images must be interpreted using a precise geometric model of the X-ray formation process. The current method treats the C-arm's intrinsic parameters (focal length, principal point, skew, and aspect ratio) as fixed, phantom-calibrated constants, while re-estimating extrinsic pose for each acquisition to align images with patient anatomy or preoperative models.

Phantom-based calibration techniques are still the clinical gold standard for achieving submillimeter accuracy. Classical algorithms like Tsai's method\cite{tsai1987versatile} and its extensions using plane-based radiopaque markers\cite{noo2004image, strobel2010camera} or checkerboard phantoms\cite{zhang2000flexible} allow precise recovery of both intrinsic and extrinsic parameters. However, these methods have a significant workflow overhead: dedicated calibration sessions, precise phantom placement and handling, and sensitivity to mechanical flex, component aging, or thermal drift.\cite{jud2018reconstruction} In crowded operating rooms or resource-constrained settings, such stringent intrinsic calibration may be impractical, delaying procedures or diverting personnel away from patient care.

In practice, many intraoperative imaging tasks rely primarily on updating the extrinsic pose (i.e., the C-arm's position and orientation relative to the patient or surgical space) while maintaining fixed intrinsic parameters obtained from a prior calibration session. Although extrinsic re-estimation can partially compensate for systematic imaging changes, this approach rests on the assumption that the intrinsic calibration remains accurate over time. Unfortunately, this assumption frequently breaks down: C-arm systems may undergo mechanical reconfiguration, thermal expansion, or cumulative component wear, all of which can introduce gradual or sudden shifts in intrinsic parameters. When the assumed intrinsic model becomes outdated or mismatched, reconstruction quality may degrade, potentially compromising clinical decision-making.

This leads to a fundamental question: \textit{How much intrinsic calibration inaccuracy can be tolerated before 3D reconstruction performance suffers significantly?} Understanding the sensitivity of reconstruction accuracy to intrinsic perturbations is critical for establishing acceptable calibration tolerances and assessing whether reliance on outdated or approximate intrinsic parameters remains clinically viable. If moderate intrinsic errors can be effectively mitigated by extrinsic re-optimization, then a less stringent calibration protocol may be sufficient—simplifying clinical workflows, reducing setup time, and lowering operational complexity without sacrificing accuracy.

Several recent studies have addressed related aspects of calibration robustness and error propagation in X-ray imaging. Klein et al.\cite{klein2004robust} and Strobel et al.\cite{strobel2010camera} advanced phantom-based techniques by modeling geometric distortions and employing nonlinear optimization to improve camera model accuracy. However, these methods still rely on tightly controlled calibration procedures and do not directly address how the reconstruction pipeline responds to residual intrinsic errors that persist in clinical use.

Recent work by Berger et al.\cite{berger2016marker} and Bier et al.\cite{bier2018x} has explored marker-free and learning-based calibration techniques, aiming to reduce reliance on physical phantoms. While promising, these methods still assume that intrinsic parameters must be precisely estimated before extrinsic registration can proceed reliably. Little research has directly quantified the \textit{trade-off} between intrinsic accuracy and reconstruction fidelity under real-world conditions, or systematically measured how extrinsic-only compensation can tolerate moderate intrinsic errors.

In this study, we take a different approach: instead of improving intrinsic calibration techniques, we assess \textit{how much intrinsic inaccuracy the system can tolerate} if we rely solely on extrinsic refinement to compensate. By systematically introducing controlled perturbations to focal length and principal point and observing their impact on 3D reconstruction and reprojection errors, we establish practical bounds on calibration robustness. Our simulation and experimental results demonstrate that submillimeter 3D reconstruction accuracy can be sustained even with intrinsic errors exceeding 100 mm in physical focal length, provided extrinsic parameters are recomputed adaptively.

The contributions of this work are threefold:
\begin{enumerate}
    \item We provide a systematic characterization of 3D reconstruction robustness to intrinsic calibration errors, spanning a wide range of perturbations in focal length and principal point.
    
    \item We validate our findings through both simulation and real-world experiments on five independent C-arm systems, demonstrating that moderate intrinsic inaccuracies can be effectively mitigated by extrinsic-only re-optimization.
    
    \item We establish practical calibration tolerance bounds that support a more flexible, clinically efficient workflow—reducing the need for frequent or highly precise intrinsic recalibration.
\end{enumerate}

Our results show that extrinsic-only compensation is remarkably robust, with submillimeter accuracy maintained under intrinsic perturbations far exceeding typical clinical noise. These findings have important implications for image-guided surgery, intraoperative navigation, and quantitative C-arm imaging—suggesting that simplified calibration procedures may be sufficient for most applications, thereby reducing operational burden without sacrificing accuracy.

\section{METHODS}

\subsection{Simulation Environment Setup}
\label{sec:simulation}

We developed a simulation environment to systematically evaluate the impact of intrinsic calibration errors on 3D reconstruction accuracy in C-arm imaging systems. The image formation process was modeled using the standard pinhole camera model, which assumes a linear projection from 3D world coordinates to 2D image coordinates via a central projection through the optical center. For simplicity, we excluded lens distortion effects—such as radial and tangential distortion—focusing solely on the influence of intrinsic parameters (e.g., focal length and principal point) within an idealized, distortion-free setting. This assumption isolates the geometric contribution of intrinsic perturbations and reflects common practice when distortion is either minimal or corrected in preprocessing. Ground-truth extrinsic parameters were obtained from our prior calibrated C-arm configurations, simulating two orthogonal imaging poses—anterior-posterior (AP) and lateral (LAT)—to mimic a typical real biplanar fluoroscopy setup. Due to the non-isocentric design of most C-arm systems—which leads to slightly different source-to-detector distances and projection geometries between views—we assigned distinct focal lengths to each orientation: 4500 pixels for the AP view and 4550 pixels for the LAT view. These values closely approximate the ground truth for 9-inch C-arm detectors based on empirical observations. The principal points were positioned at the center of $1024 \times 1024$ images (with a pixel spacing of 0.21 mm/pixel), consistent with the resolution of high-end clinical imaging systems.

\subsubsection{Intrinsic Perturbation Modeling}

To replicate real-world calibration imperfections, we introduced controlled perturbations to the intrinsic parameters—focal length and principal point—by adding random noise sampled from a uniform distribution in the range $[-1, 1]$, scaled separately for each parameter. This approach models typical deviations arising from suboptimal phantom-based calibration, environmental influences, or device-specific variability.

For each perturbed intrinsic configuration, extrinsic parameters were re-estimated using synthetic 2D projections of known 3D landmarks and a Perspective-n-Point (PnP) algorithm.\cite{lepetit2009epnp} This mimics clinical scenarios where extrinsics are refined based on limited image-based cues, assuming inaccurate or outdated intrinsic calibrations.

\subsubsection{Evaluation Procedure}

To assess reconstruction robustness under intrinsic uncertainty, we conducted a Monte Carlo analysis across multiple perturbation levels. Each trial involved the following steps:
\begin{itemize}
    \item \textbf{Projection:} 3D points were projected into both views using perturbed intrinsics and estimated extrinsics.
    \item \textbf{Triangulation:} 3D points were reconstructed using linear triangulation from 2D correspondences.
    \item \textbf{Alignment \& Error Computation:} The reconstructed point cloud was aligned to the ground-truth using rigid Procrustes analysis,\cite{gower2004procrustes} and root mean square error (RMSE) was computed to measure reconstruction accuracy. Reprojection errors were also calculated by comparing observed and back-projected 2D points.
\end{itemize}

\begin{figure}[ht]
\centering
\includegraphics[width=\linewidth]{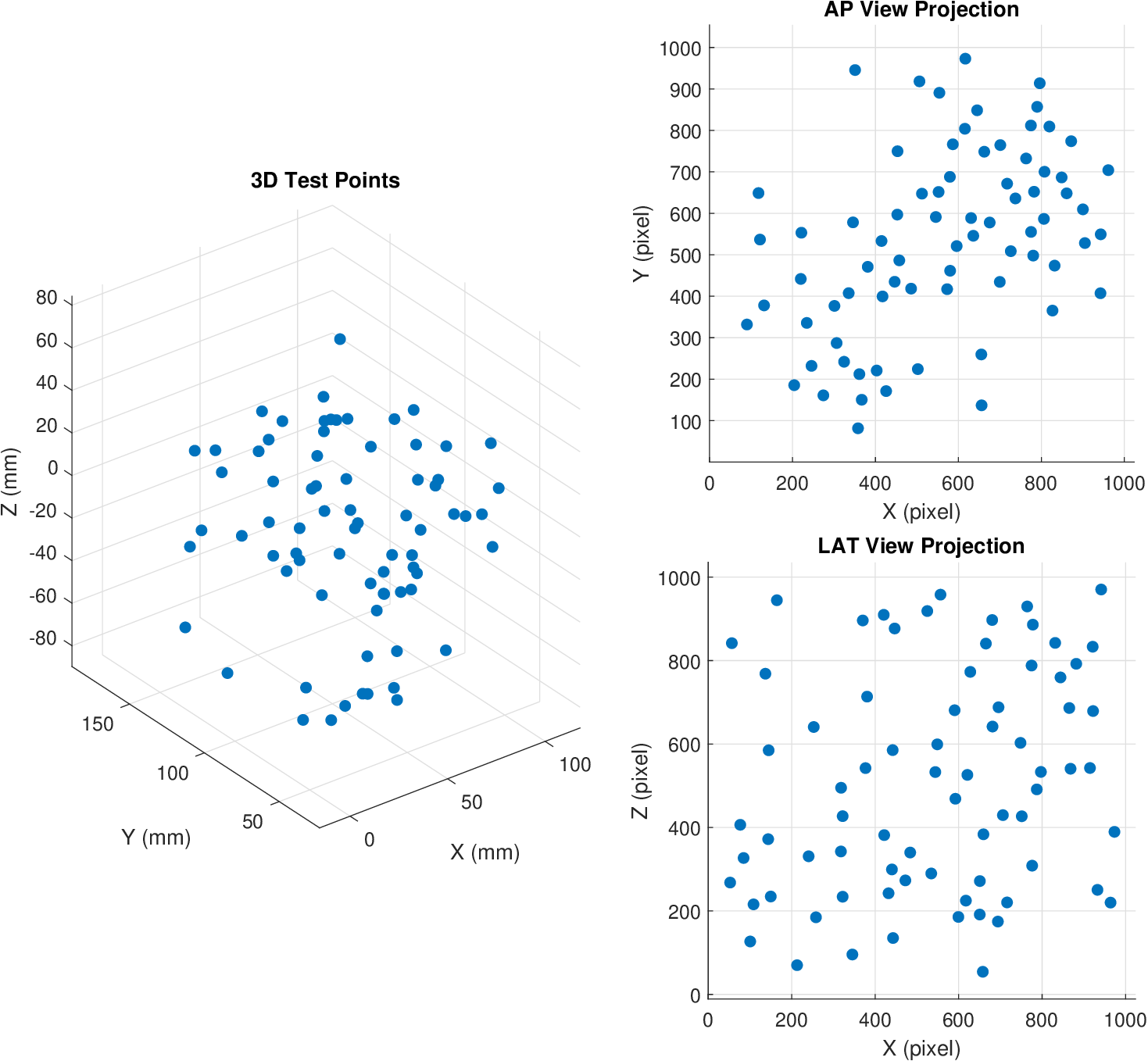}
\caption{Visualization of 3D test points and their corresponding 2D projections in AP and LAT views. \textbf{Left:} A total of 74 3D test points selected from an initial set of 500 uniformly distributed samples within the overlapping field of view of the AP and LAT imaging planes. These points are located within the region where both views provide valid projection coverage, ensuring reliable 3D reconstruction. \textbf{Top right:} The 2D projections of the selected 3D points onto the AP image plane, simulating how the C-arm captures these spatial locations in the anterior-posterior view. \textbf{Bottom right:} The corresponding 2D projections of the same 3D points onto the LAT image plane, representing the lateral perspective projection.}
\label{fig:simulated_rand_points}
\end{figure}

\subsubsection{Test Volume and Point Sampling}

We defined a 150 $\times$ 150 $\times$ 150 mm$^3$ test volume that covers the overlapping field of view of the AP and LAT projections, representing a clinically relevant region for accurate 3D reconstruction. This volume was carefully selected to ensure that it lies entirely within the shared visibility cone of both imaging views, thus allowing robust triangulation while avoiding projection overlap and occlusion effects near the image periphery. To ensure comprehensive coverage of the volume, 500 random 3D points were uniformly sampled, from which 74 were selected based on the following filtering criteria:
\begin{itemize}
    \item \textbf{Edge Exclusion:} Points that projected too close to the image boundaries (with a root mean square (RMS) distance less than 40 pixels from the edge) were discarded. This threshold helps avoid instability and geometric distortion commonly found at the periphery of the image, improving the reliability of projections and visualization clarity.
    
    \item \textbf{Separation Filtering:} Points exhibiting small disparities (RMS $<$ 40 pixels) between their AP and LAT projections were excluded. A minimum disparity threshold of 40 pixels ensures sufficient stereo baseline, which is critical for accurate 3D triangulation and reduces ambiguity in depth estimation.
\end{itemize}

This sampling strategy ensured even spatial distribution throughout the volume while maintaining high numerical stability and geometric reliability in the projected views. Figure~\ref{fig:simulated_rand_points} illustrates the selected 3D landmarks and their projections in AP and LAT views.

\subsubsection{Metrics}

We evaluated system performance using:
\begin{itemize}
    \item \textbf{3D Reconstruction Error:} RMSE of the Euclidean distance between reconstructed and ground-truth 3D points after alignment, reported in millimeters (mm).
    
    \item \textbf{Reprojection Error:} Average pixel distance between the observed 2D image points and the corresponding points reprojected from the reconstructed 3D positions, reported in pixels.
\end{itemize}

By varying the magnitude of intrinsic perturbations across Monte Carlo trials, this framework provides a rigorous characterization of how intrinsic inaccuracies influence both extrinsic estimation and 3D reconstruction accuracy.

\subsection{Real C-arm Data}

To validate the robustness of our method in practical settings, we conducted experiments on five commercially available GE OEC 9900 Elite 9-inch C-arm systems (serial numbers: E0284, E1272, E1938, E1090, and E2706). These systems were evaluated in a typical clinical environment to assess how intrinsic miscalibrations affect 3D reconstruction performance.

\subsubsection{Baseline Calibration}

Each C-arm was first calibrated using a standard phantom-based technique\cite{zhang2000flexible} to obtain a baseline estimate of the intrinsic camera matrix $K$. This procedure reflects the initial setup used in clinical workflows, where a known calibration phantom (e.g., checkerboard or radiopaque markers) is used to compute the focal lengths and principal point locations for the AP and LAT views. The resulting matrix served as the ground-truth or reference configuration for subsequent perturbation experiments.

\subsubsection{Perturbation of Intrinsic Parameters}

To simulate realistic intrinsic calibration errors, we introduced controlled perturbations to the intrinsic matrix $K$ of each C-arm system. Specifically, we modified the focal lengths and principal point coordinates to mimic common sources of error, including:
\begin{itemize}
    \item Limited phantom coverage during calibration,
    \item Poor image quality due to noise or motion blur affecting the BBs identification,
    \item Inconsistent system setup across imaging sessions.
\end{itemize}

Perturbations were applied independently for the AP and LAT views to reflect the asymmetric geometry often observed in clinical biplanar imaging setups. The magnitude of these perturbations was informed by empirical calibration logs from clinical environments and aligned with the noise levels used in our simulation studies.

We systematically varied the intrinsic parameters by introducing focal length deviations ranging from $\pm 100$ to $\pm 500$ pixels and principal point offsets between $\pm 20$ and $\pm 200$ pixels. Based on our experiments and prior knowledge, this range represents the most common levels of calibration noise. For each perturbed configuration, extrinsic parameters were re-estimated using a Perspective-n-Point (PnP) algorithm.

To evaluate reconstruction robustness, we used 32 predefined test points from a known calibration phantom. Experiments were conducted across five different GE OEC 9900 Elite 9-inch C-arm systems at various anatomical locations and time points, ensuring variability and generalizability of the results.

\begin{figure}[ht]
\centering
\includegraphics[width=13cm]{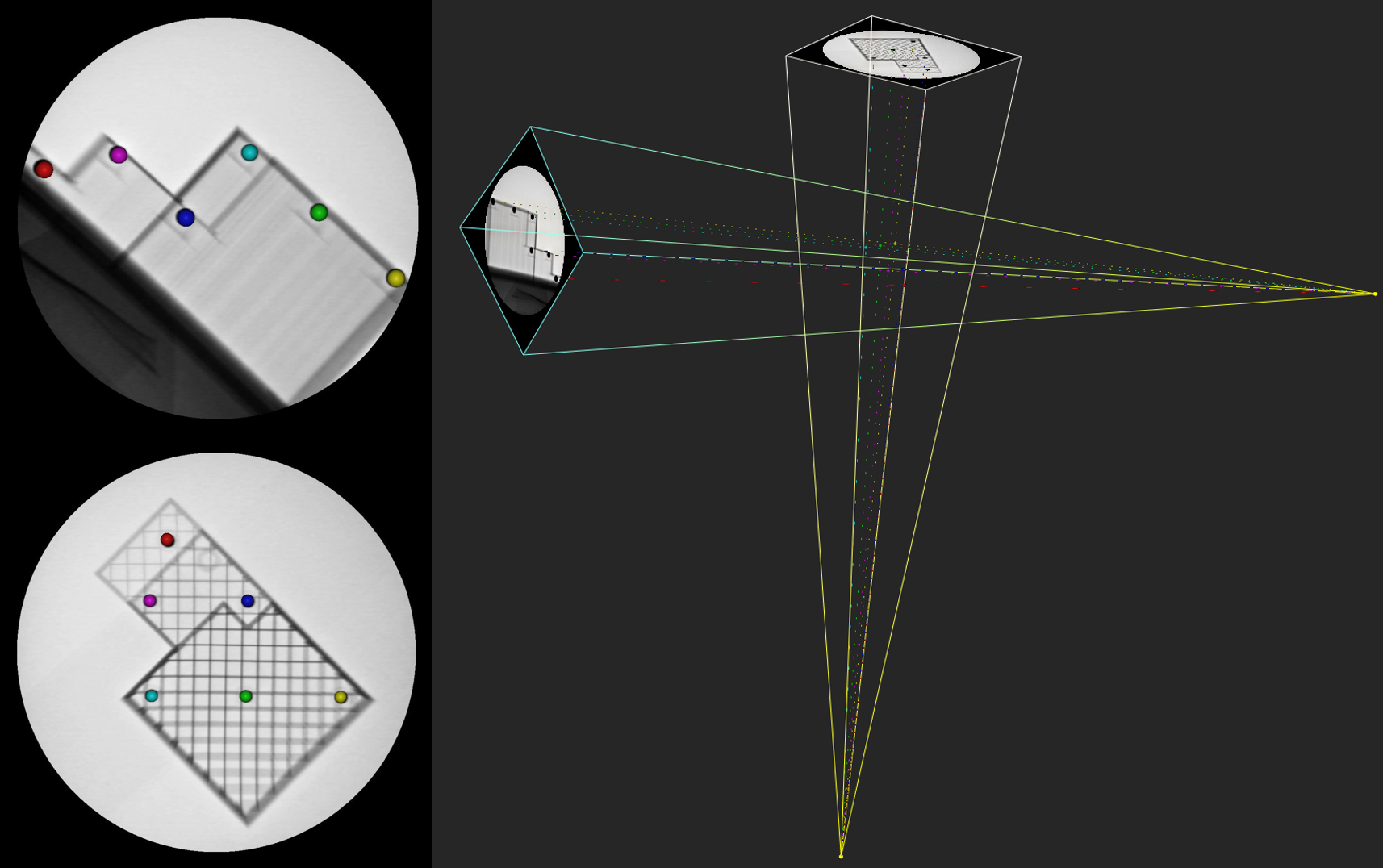}
\caption{3D environment constructed from real C-arm data for extrinsic parameter re-estimation: AP and LAT images of the phantom with back-projected reconstructed BBs (left), and a 3D view of the scene with projection lines visualized (right).}
\label{fig:real_env3D}
\end{figure}

\subsubsection{Image Acquisition}

For each perturbed intrinsic configuration, two orthogonal X-ray images—AP and LAT views—were acquired using a clinical 3D calibration phantom consisting of radiopaque markers arranged in a known 3D configuration. To ensure robustness under variable conditions, we acquired four independent pairs of AP and LAT images per C-arm system. These image pairs were collected from different spatial locations (i.e., varying C-arm positions and orientations) and at different times, reflecting common use-case variability in surgical or interventional workflows.

\subsubsection{Extrinsic Re-estimation and Reconstruction}

Given the perturbed intrinsic matrix and acquired image pairs, we re-estimated the extrinsic parameters (rotation and translation matrices) using a Perspective-n-Point (PnP) algorithm. The re-estimation process minimized the reprojection error between the observed 2D image points and the known 3D positions of the phantom markers, emulating standard image-based registration techniques used intraoperatively.

With the perturbed intrinsics and re-estimated extrinsics, we performed 3D reconstruction of the phantom marker positions via linear triangulation from the AP and LAT projections. This allowed us to evaluate the sensitivity of the full reconstruction pipeline—calibration, registration, and triangulation—to errors in intrinsic calibration.

\subsubsection{Evaluation and Visual Inspection}

To assess reconstruction accuracy, the 3D reconstructed point cloud was rigidly aligned to the ground-truth marker geometry using rigid Procrustes analysis, and the resulting error was quantified using RMSE. Visual overlays of the projections were also generated to assess consistency across views.

Figure~\ref{fig:real_env3D} shows a representative example of the reconstructed phantom in 3D, along with the corresponding AP and LAT images used in the 3D reconstruction.

Since the true intrinsic and extrinsic parameters of the clinical C-arm system are not directly observable, we evaluated the reconstruction accuracy by comparing the estimated 3D landmark positions to the known geometry of the calibration phantom. This phantom served as the ground-truth reference in physical space. The primary evaluation metric was the 3D reconstruction error, computed as the Root Mean Square Error (RMSE) between the reconstructed 3D points and the corresponding known ground-truth positions. All errors were reported in millimeters, reflecting the geometric fidelity of the reconstruction under varying intrinsic perturbations. This approach provides a practical and clinically relevant measure of reconstruction robustness in the absence of directly measurable ground-truth system parameters.

\section{RESULTS}

In this section, we present a comprehensive evaluation of the proposed calibration-robust 3D reconstruction framework, using both simulated and real-world data. We begin by analyzing the sensitivity of reconstruction accuracy to controlled intrinsic perturbations in a synthetic biplanar imaging environment. This simulation study allows us to quantify how errors in focal length and principal point influence final 3D accuracy across a range of noise scales.

We then validate our findings using real data acquired from five commercial C-arm systems. By applying the same perturbation and extrinsic re-estimation procedure to clinical X-ray images of a known phantom, we assess the robustness of the approach under realistic imaging conditions. Together, these two parts establish a foundation for understanding the practical implications and reliability of bypassing conventional intrinsic recalibration made to usually maintain accuracy in intrinsic parameters.

\subsection{Simulated Data}

To assess the robustness of the 3D reconstruction under intrinsic perturbation, we evaluated reconstruction error across a range of intrinsic variations. Specifically, we simulated perturbations in focal length (±100 to ±700 pixels) and principal point offsets (±20 to ±200 pixels), and computed the corresponding extrinsic parameters through PnP. Reconstruction accuracy was assessed from 74 randomly sampled test points within the 150 mm$^3$ volume, filtered as described in Section~\ref{sec:simulation}.

\begin{figure}[ht]
\centering
\includegraphics[width=0.45\textwidth]{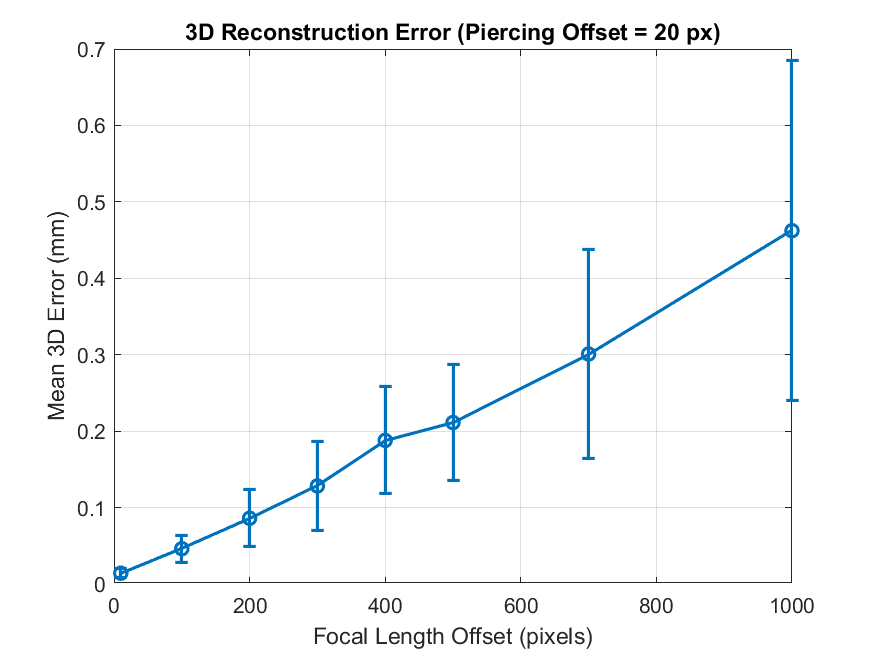}
\hfill
\includegraphics[width=0.45\textwidth]{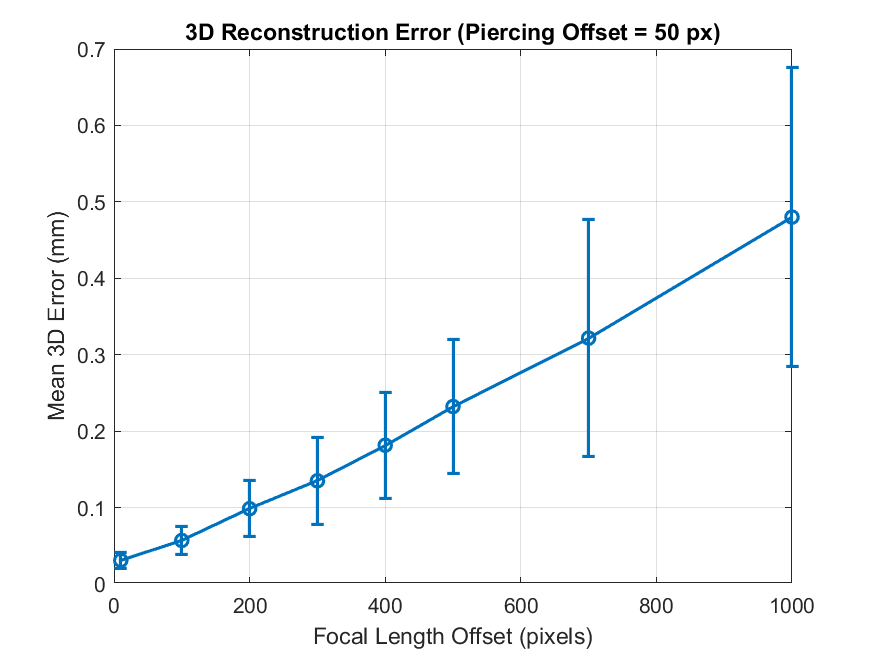}

\vspace{0.5em}

\includegraphics[width=0.45\textwidth]{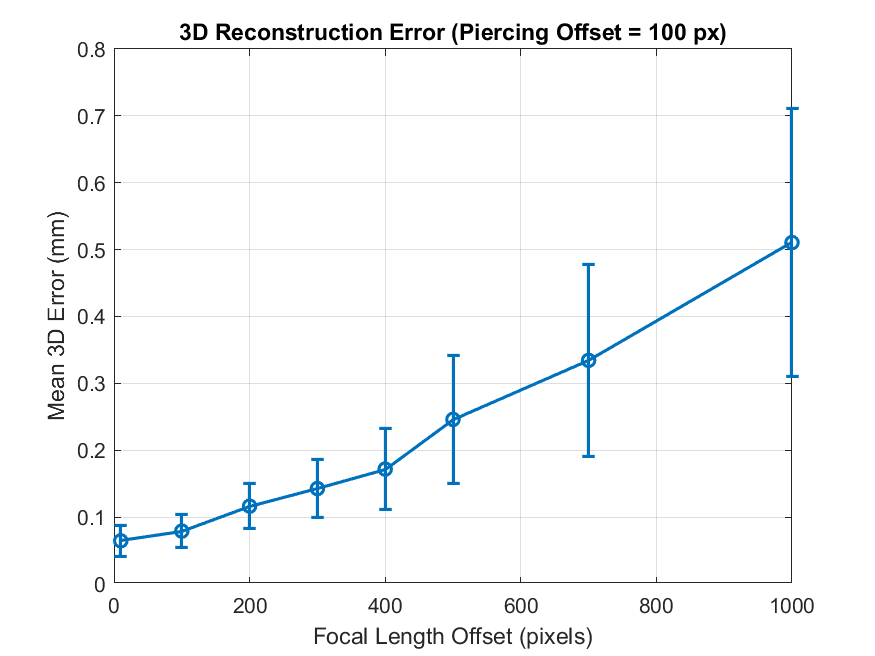}
\hfill
\includegraphics[width=0.45\textwidth]{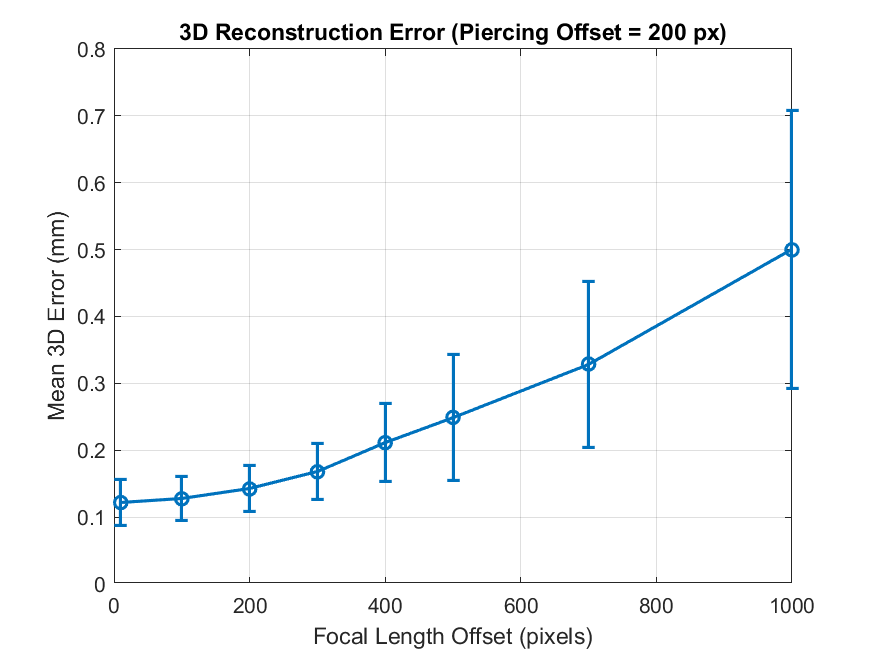}

\caption{Mean and standard deviation of the 3D reconstruction error versus focal length perturbation for piercing point offsets of 20 px (top left), 50 px (top right), 100 px (bottom left), and 200 px (bottom right). Errors remain below 0.5 mm even with ±700 px focal length shifts.}
\label{fig:recon_error_grid}
\end{figure}

Figure~\ref{fig:recon_error_grid} illustrates the relationship between focal length perturbation and 3D reconstruction accuracy under varying levels of principal point (piercing point) offsets. Across all scenarios, the reconstruction error remained below 0.5 mm, even for focal length deviations as large as ±700 pixels. Given the system resolution of 0.21 mm per pixel, this corresponds to a focal length perturbation of approximately 140 mm—yet the reconstruction performance remained consistently robust.

In addition, we evaluated the sensitivity of reconstruction accuracy to principal point deviations by introducing offsets of 20, 50, 100, and 200 pixels. As the principal point offset increased from 20 to 200 pixels, the 3D reconstruction error increased by less than 0.05 mm across all focal length perturbation levels.

\begin{figure}[ht]
\centering
\includegraphics[width=0.45\textwidth]{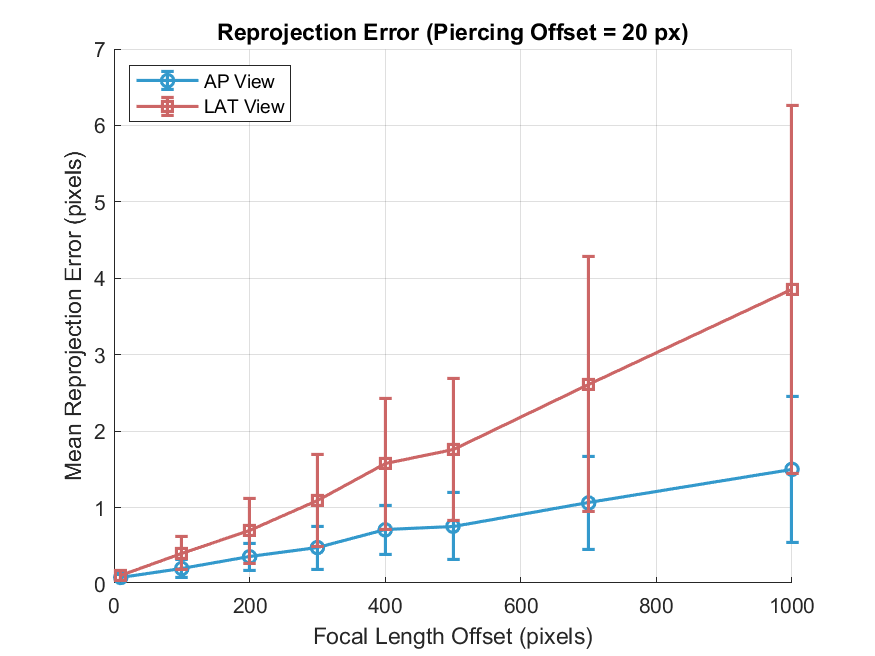}
\hfill
\includegraphics[width=0.45\textwidth]{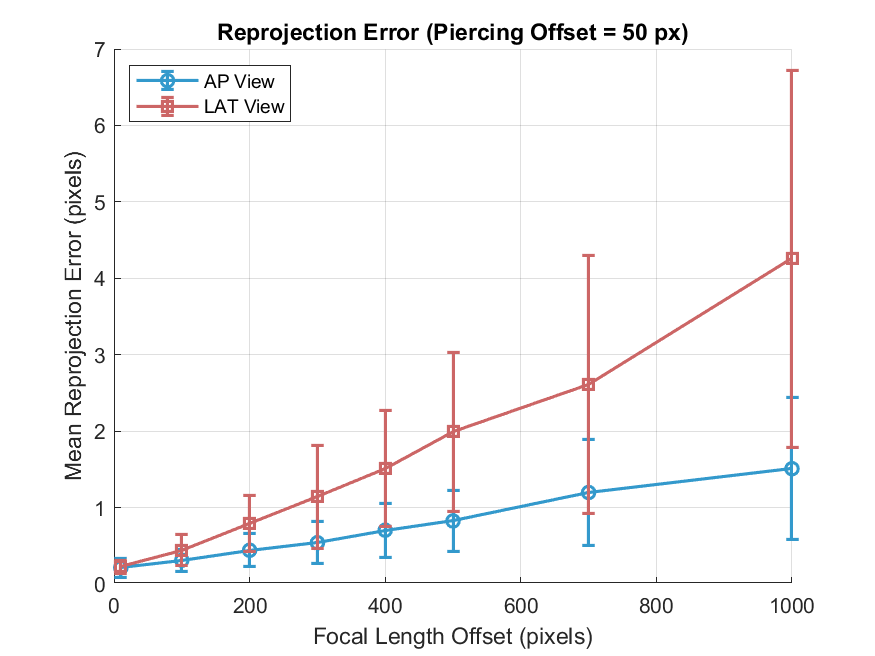}

\vspace{0.5em}

\includegraphics[width=0.45\textwidth]{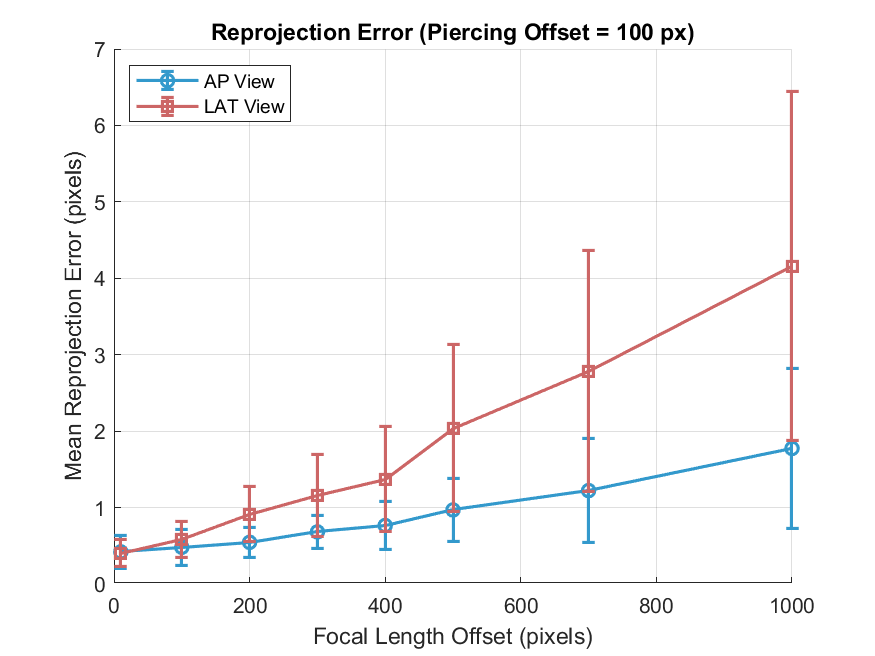}
\hfill
\includegraphics[width=0.45\textwidth]{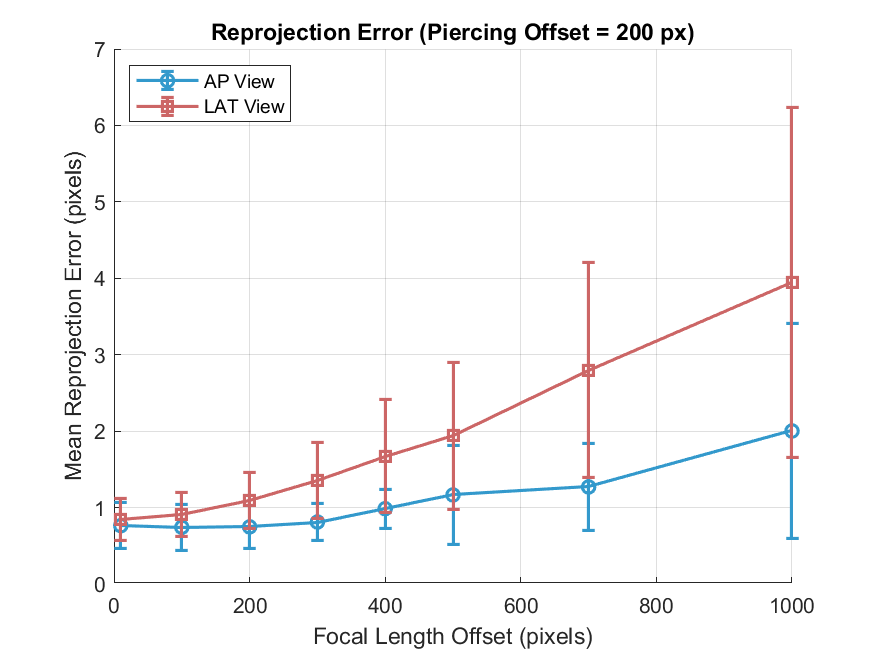}

\caption{Mean and standard deviation of the reprojection error as a function of focal length perturbation, shown for piercing point offsets of 20 px (top left), 50 px (top right), 100 px (bottom left), and 200 px (bottom right). In all cases, the reprojection error remains below 5 pixels for both AP and LAT views, even with focal length shifts up to ±700 px.}
\label{fig:combine_reproject_error_simulated}
\end{figure}

Figure~\ref{fig:combine_reproject_error_simulated} presents the mean and standard deviation of the reprojection error as a function of focal length perturbation magnitude. Despite increasing deviations in focal length, the system consistently maintained reprojection errors below 5 pixels in both the anterior-posterior (AP) and lateral (LAT) views, even with perturbations up to ±700 pixels. Given the system's spatial resolution of 0.21 mm per pixel, this corresponds to a maximum mean reprojection error of less than 1 mm—demonstrating the robustness of the extrinsic compensation strategy.

We also investigated the effect of principal point (piercing point) perturbations at four different offsets: 20, 50, 100, and 200 pixels. As the principal point offset increased from 20 to 200 pixels, the reprojection residual error increased by less than 0.5 pixel across all focal length perturbation levels either in AP or LAT view.

\subsection{Real Data Results and Analysis}

\begin{figure}[ht]
\centering
\includegraphics[width=0.45\textwidth]{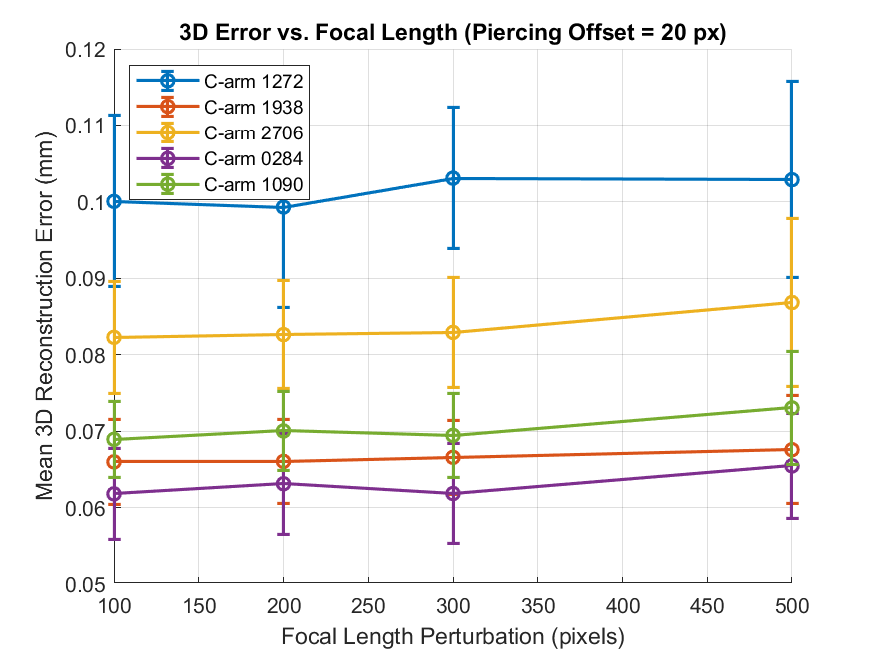}
\hfill
\includegraphics[width=0.45\textwidth]{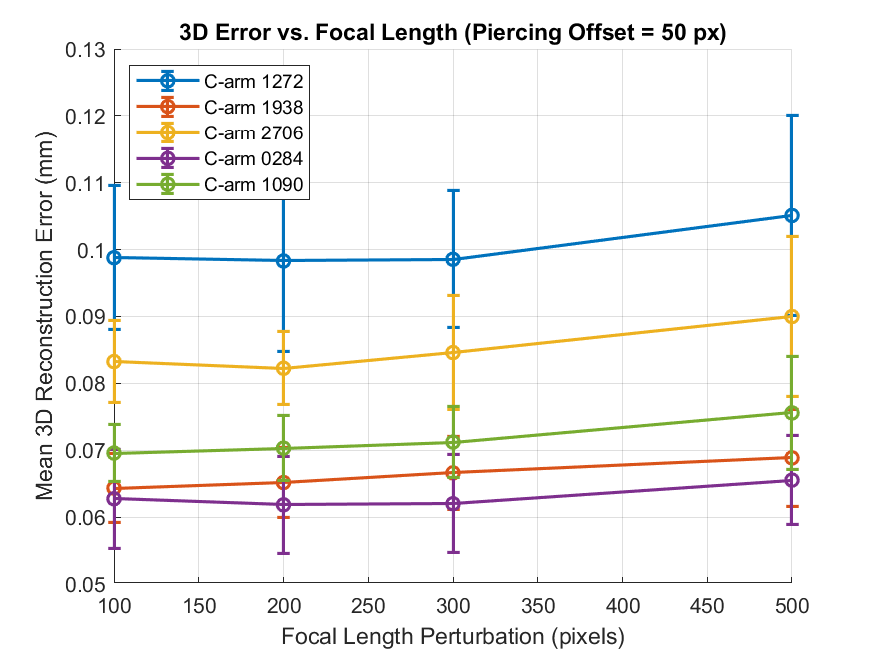}

\vspace{0.5em}

\includegraphics[width=0.45\textwidth]{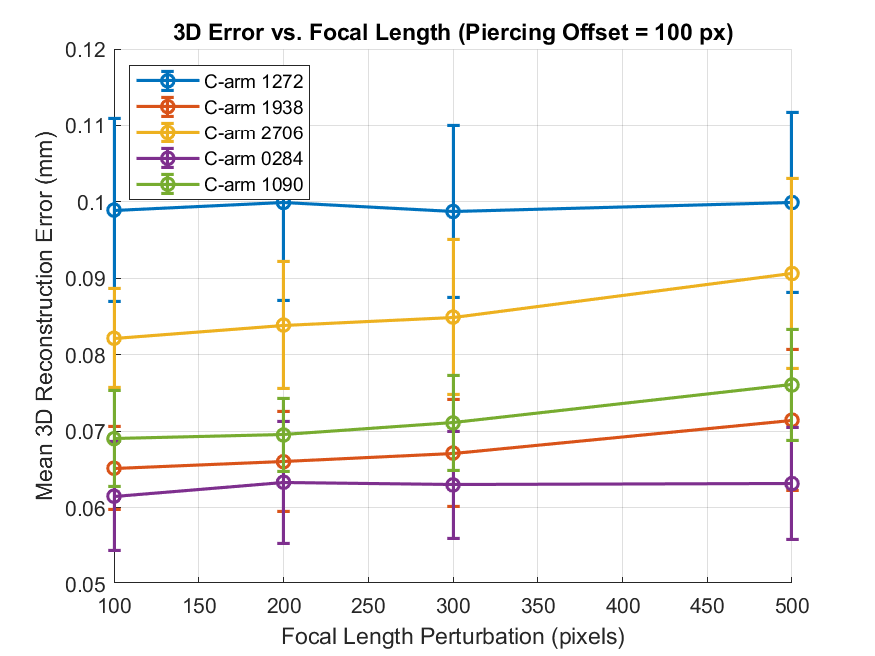}
\hfill
\includegraphics[width=0.45\textwidth]{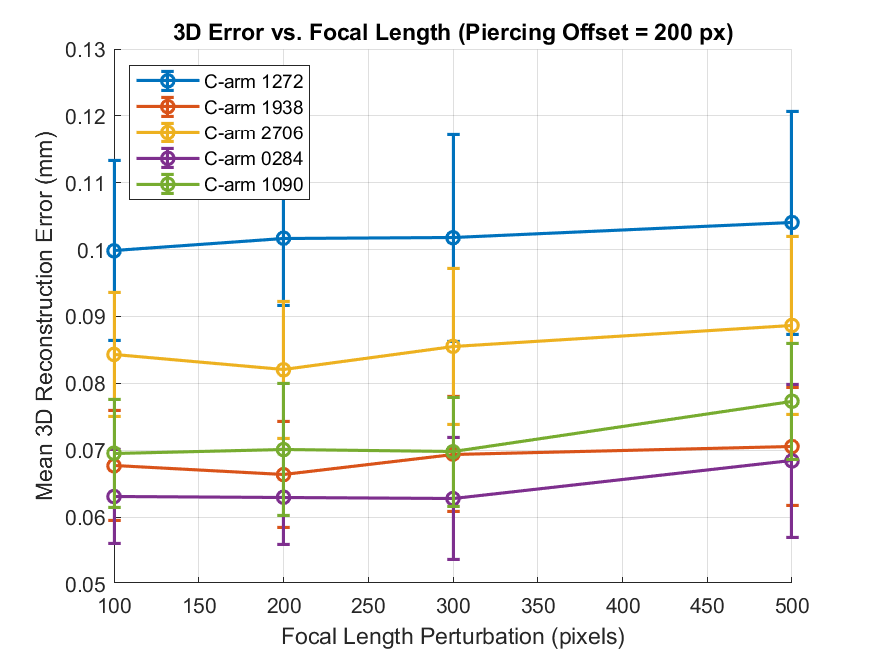}

\caption{Mean and standard deviation of the 3D reconstruction error as a function of focal length perturbation, shown for piercing point offsets of 20 px (top left), 50 px (top right), 100 px (bottom left), and 200 px (bottom right). Across all settings, errors remain below 0.2 mm even with focal length shifts up to ±500 px.}
\label{fig:3d_reconstruct_real_data}
\end{figure}

Figure~\ref{fig:3d_reconstruct_real_data} presents the mean and standard deviation of the 3D reconstruction error as a function of focal length perturbation. Across all five C-arm systems evaluated, the reconstruction consistently achieved submillimeter accuracy, with mean errors remaining below 0.2 mm—even when focal length deviations reached up to ±500 pixels. Given the system's imaging resolution of 0.21 mm per pixel, this level of deviation corresponds to approximately ±100 mm of physical focal length error, assuming a nominal focal length of $\sim$1000 mm. Despite such perturbations, the reconstruction performance remained stable and precise. The ground-truth focal lengths used in our experiments were 4800 pixels for the AP view and 4850 pixels for the LAT view, serving as the baseline for all focal length perturbation tests.

Performance trends were consistent across systems. For example, C-arm E0284 achieved the highest accuracy, with mean reconstruction errors below 0.08 mm. In contrast, C-arm E1272 produced the largest observed error, reaching approximately 0.13 mm.

To further evaluate sensitivity to intrinsic calibration inaccuracies, we introduced synthetic principal point offsets of 20, 50, 100, and 200 pixels. Across all perturbation levels and C-arm systems, the resulting 3D reconstruction errors remained below 0.01 mm, indicating strong robustness to principal point misalignment.

\section{DISCUSSION AND CONCLUSION}

\subsection{Discussion}

This study analyzed how intrinsic calibration errors—specifically deviations in focal length and principal point—affect 3D reconstruction accuracy in C-arm–based imaging. Using both simulation and experimental validation on clinical systems, we demonstrated that these intrinsic errors can be effectively mitigated by re-estimating the extrinsic parameters alone, without requiring precise intrinsic recalibration.

In simulation, we introduced controlled perturbations of up to ±700 pixels in focal length and ±200 pixels in principal point to emulate realistic calibration imperfections. Across these perturbations, both 3D reconstruction and reprojection errors remained remarkably stable, consistently achieving submillimeter accuracy. Re-optimizing the extrinsic pose via standard Perspective-n-Point (PnP) and bundle adjustment methods was sufficient to compensate for even substantial intrinsic deviations. Importantly, the influence of principal point shifts was minimal, suggesting that focal length variation plays a dominant role in reconstruction stability.

One notable observation was the consistently higher reprojection error in the lateral (LAT) view compared to the anterior-posterior (AP) view. We attribute this to the LAT view typically being positioned closer to the object, which increases the sensitivity of projection accuracy to small pose estimation errors—a known limitation of PnP-based methods in short baseline or high-zoom geometries. While reconstruction quality remained within acceptable limits (1--2 mm),\cite{Wein2008, Uneri2013, MaierHein2013} this directional asymmetry highlights an area for future investigation, especially in optimizing multi-view fusion strategies.

These findings were further validated through experiments on five GE OEC 9900 C-arm systems using phantom and anatomical targets. All systems demonstrated submillimeter 3D reconstruction accuracy, with the best case in average achieving 0.08 mm error and the worst remaining below 0.13 mm. This consistent performance across independent hardware units and acquisition conditions reinforces the robustness and generalizability of the extrinsic-only compensation approach.

From a clinical perspective, our results show that extrinsic refinement alone (using fixed pre-calibrated intrinsic) can preserve quantitative accuracy—even under significant intrinsic perturbations—this study supports a more flexible, scalable calibration workflow. This has practical implications for streamlining intraoperative imaging in resource-constrained or time-critical settings, where repeated phantom-based calibration may be infeasible. Prior work suggests that 3D accuracy within 1--2 mm is sufficient for most image-guided procedures, and our method remains well within this threshold under worst-case intrinsic deviations.

In summary, we provide a comprehensive analysis of how intrinsic parameter inaccuracies propagate through the 3D reconstruction pipeline. By leveraging extrinsic re-optimization, we show that robust geometric accuracy can be maintained without strong dependence on precise intrinsic calibration—enabling more practical and resilient imaging workflows for surgical navigation and quantitative C-arm applications.

\subsection{Limitations and Future Work}

While this study demonstrates the robustness of extrinsic compensation under intrinsic perturbations, it assumes a simplified perspective projection model with minimal or corrected geometric distortion. In practice, C-arm systems with significant lens or geometric distortion may still require explicit distortion modeling or conventional intrinsic calibration to ensure accurate 3D reconstruction. Additionally, our evaluation was limited to stereo image pairs (antero-posterior and lateral views), and it remains uncertain whether the observed robustness generalizes to more complex, multi-view configurations—an important consideration for broader 3D reconstruction workflows such as those used in cone-beam CT or mobile fluoroscopy systems. While our work focuses on structural 3D reconstruction accuracy, emerging applications 
in dynamic medical imaging \cite{wang2024deep, wang2021functional} may benefit from similar robustness analysis of calibration-dependent imaging pipelines.

Moreover, the current simulation pipeline relies on random sampling strategies to generate intrinsic perturbations. While effective, these may not fully capture the variance in clinical calibration scenarios. As part of future work, we plan to incorporate quasi-Monte Carlo (QMC) sampling methods\cite{niederreiter1992random, Li2020_QMC, LiHyman2025_RBF_CVI} to achieve more uniform and efficient coverage of the perturbation space. QMC has been shown to improve the convergence rate and statistical efficiency of uncertainty quantification in medical imaging applications\cite{kaiser2019qmc, lee2022qmc} and may enhance the robustness and generalizability of our simulation-based validation framework.

Future work will also extend this framework to more complex scenarios, such as 3D model reconstruction, that involve including multi-angle imaging (e.g., five or more views) to ensure 3D model reconstruction fidelity. Additionally, we aim to integrate learning-based registration or pose estimation techniques to enable real-time extrinsic optimization, particularly in intraoperative workflows where speed and robustness are critical. Finally, clinical validation in surgical environments will be essential to evaluate the practical effectiveness, limitations, and scalability of the proposed strategy in real-world settings.

\subsection{Conclusion}

This work redefines the importance of precise intrinsic calibration in C-arm–based 3D reconstruction by demonstrating that high reconstruction accuracy can be preserved even in the presence of substantial intrinsic parameter errors. We show that deviations in focal length and principal point—within ranges commonly encountered in clinical environments—can be effectively compensated through extrinsic re-optimization. By systematically quantifying 3D reconstruction and reprojection errors across a wide range of intrinsic perturbations, we establish practical bounds on calibration robustness. These results are particularly relevant for medical applications such as image-guided surgery, intraoperative navigation, and 3D anatomical localization, where frequent recalibration is impractical. Our findings support a more flexible and clinically efficient calibration workflow, enabling robust C-arm–based 3D reconstruction without the need for tightly constrained intrinsic precision.

\section*{ACKNOWLEDGMENTS}

This work was supported by a study grant from Alphatec Spine Inc. The sponsor had no involvement in the study design, data collection, analysis, interpretation, or manuscript preparation.

\section*{CONFLICT OF INTEREST}

All authors are full-time employees of Alphatec Spine, Inc., which may be considered a potential competing interest.

\bibliography{bibliography}
\bibliographystyle{spiebib}

\end{document}